\documentclass[runningheads]{llncs}
\usepackage{graphicx}
\usepackage{tikz}
\usepackage{comment}
\usepackage{amsmath,amssymb} % define this before the line numbering.
\usepackage{color}

\usepackage[accsupp]{axessibility}  % Improves PDF readability for those with disabilities.

\usepackage{graphicx}
\usepackage{subfigure}
\usepackage{multirow}
\usepackage{xcolor}
\usepackage{booktabs} 

\usepackage{adjustbox}

\usepackage{floatrow}

\newcommand{\ie}{\textit{i}.\textit{e}., }

\begin{document}
% \linenumbers
\pagestyle{headings}
\mainmatter
\def\ECCVSubNumber{46}  % Insert your submission number here

\title{PPT: token-Pruned Pose Transformer for monocular and multi-view human pose estimation} % Replace with your title

%\begin{comment}
% \titlerunning{ECCV-22 submission ID \ECCVSubNumber} 
% \authorrunning{ECCV-22 submission ID \ECCVSubNumber} 
% \author{Anonymous ECCV submission}
% \institute{Paper ID \ECCVSubNumber}
%\end{comment}
%******************

% \begin{comment}

% % If the paper title is too long for the running head, you can set
% % an abbreviated paper title here

\titlerunning{PPT: token-Pruned Pose Transformer}
\author{Haoyu Ma\textsuperscript{1}, Zhe Wang\textsuperscript{1}, Yifei Chen\textsuperscript{2}, Deying Kong\textsuperscript{1}, Liangjian Chen\textsuperscript{3}, \\ 
Xingwei Liu\textsuperscript{1}, Xiangyi Yan\textsuperscript{1}, Hao Tang\textsuperscript{3}, Xiaohui Xie\textsuperscript{1}}

\authorrunning{Ma et al.}

\institute{University of California, Irvine \\
\email{\{haoyum3, zwang15, deyingk, xingweil, xiangyy4, xhx\}@uci.edu}  \and 
Tencent Inc \\
\email{dolphinchen@tencent.com} \and 
Meta Reality Lab, Meta AI \\
\email{\{clj, haotang\}@fb.com}}

% \end{comment}
%******************
\maketitle

\begin{abstract}
Recently, the vision transformer and its variants have played an increasingly important role in both monocular and multi-view human pose estimation. Considering image patches as tokens, transformers can model the global dependencies within the entire image or across images from other views. However, global attention is computationally expensive. As a consequence, it is difficult to scale up these transformer-based methods to high-resolution features and many views. 

In this paper, we propose the token-Pruned Pose Transformer (PPT) for 2D human pose estimation, which can locate a rough human mask and performs self-attention only within selected tokens. 
Furthermore, we extend our PPT to multi-view human pose estimation. Built upon PPT, we propose a new cross-view fusion strategy, called human area fusion, which considers all human foreground pixels as corresponding candidates. Experimental results on COCO and MPII demonstrate that our PPT can match the accuracy of previous pose transformer methods while reducing the computation.  Moreover, experiments on Human 3.6M and Ski-Pose demonstrate that our Multi-view PPT can efficiently fuse cues from multiple views and achieve new state-of-the-art results. 
Source code and trained model can be found at 
\url{https://github.com/HowieMa/PPT}.
% \href{https://github.com/HowieMa/TransFusion-Pose}{https://github.com/HowieMa/TransFusion-Pose}. 


\end{abstract}

\section{Introduction}

% P1: human pose estimation application, single-view and multi-view
Human pose estimation aims to localize anatomical keypoints from images. It serves as a foundation for many down-stream tasks such as AR/VR, action recognition \cite{huang2017deep,yan2018spatial}, and medical diagnosis \cite{chen2021pd}. 
Over the past decades, deep convolutional neural networks (CNNs) play a dominant role in human pose estimation tasks \cite{toshev2014deeppose,wei2016convolutional,newell2016stacked,xiao2018simple,sun2019deep,bestofboth,wang2020predicting}. 
However, cases including occlusions and oblique viewing are still too difficult to be solved from a monocular image. To this end, some works apply a multi-camera setup \cite{simon2017hand,wang2019geometric,h36m_pami,chen2021mvhm} to boost the performance of 2D pose detection\cite{qiu2019cross,he2020epipolar}, since difficult cases in one view are potentially easier to be resolved in other views. 
Meanwhile, human body joints are highly correlated, constrained by strong kinetic and physical constraints \cite{tompson2014joint}. 
However,since the reception fields of CNNs are limited,
the long-range constraints among joints are often poorly captured \cite{li2021tokenpose}.

\begin{figure}[!t]
    \centering
    \includegraphics[width=0.95\linewidth]{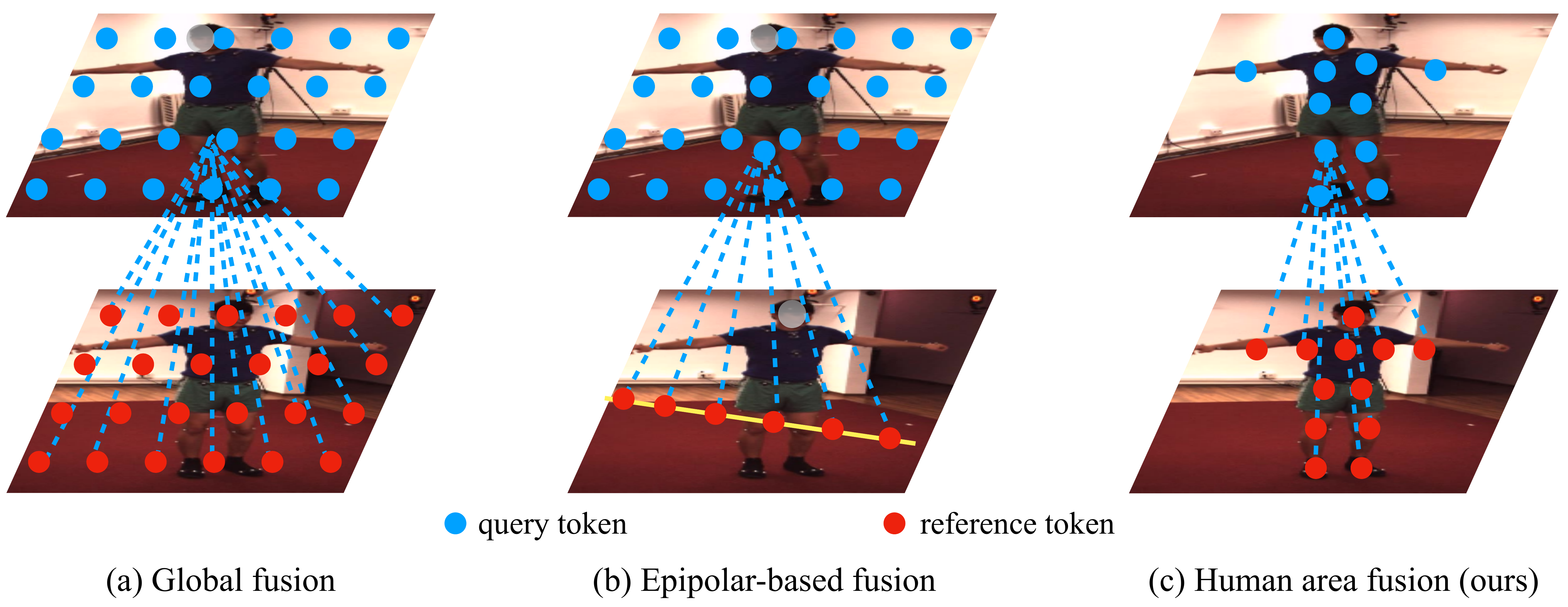}
    \caption{\small{Different types of cross-view fusion. The first row is the current view, and the second row is the reference view. }  }
    \label{fig:cross_fusion_type}
\end{figure}

\begin{figure}[t]
    \centering
    \includegraphics[width=0.88\linewidth]{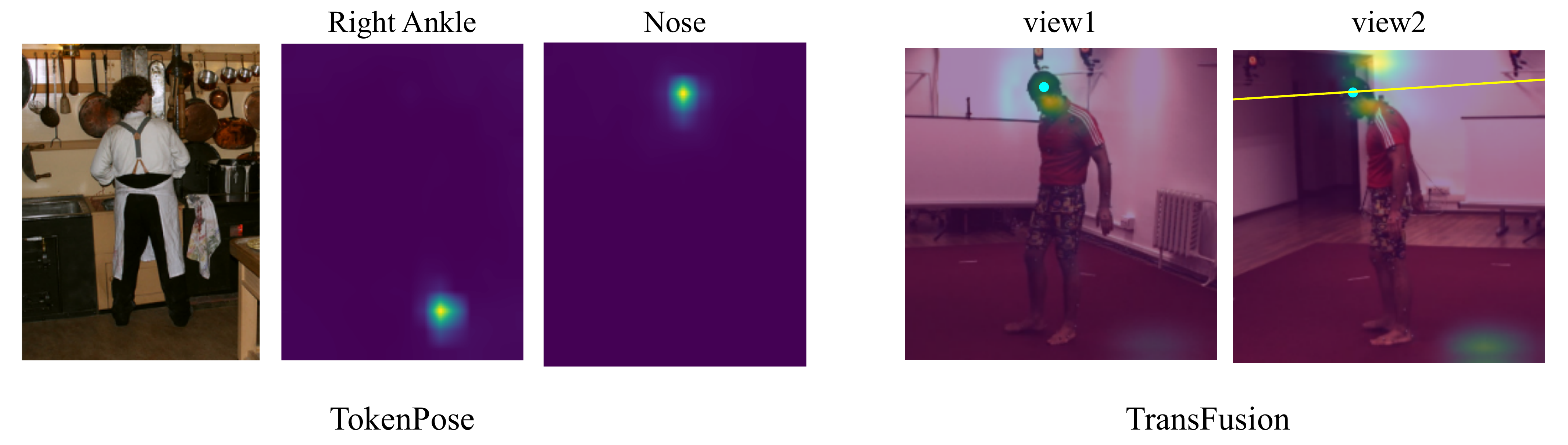}
    \caption{\small{Attention map for TokenPose (monocular view) and TransFusion (multi-view). The attention maps are very sparse and only attend to a small local regions.}}
    \label{fig:attn_map_demo}
\end{figure}

% P2: Transformer and its limitations 
Recently, the ViT \cite{dosovitskiy2020image} demonstrates that the transformers \cite{vaswani2017attention} can achieve impressive performance on many vision tasks \cite{touvron2020training,carion2020end}. 
Compared with CNN, the self-attention module of transformers can easily model the global dependencies among all visual elements. 
In the field of pose estimation, many tansformer-based works  \cite{li2021tokenpose,yang2020transpose,mao2021tfpose,lin2020end,zheng20213d} suggest that the global attention is necessary. 
In single-view 2D human pose estimation, TransPose \cite{yang2020transpose} and TokenPose \cite{li2021tokenpose} achieve new state-of-the-art performance and learn the relationship among keypoints with transformers. 
In multi-view human pose estimation, the TransFusion \cite{ma2021transfusion} uses the transformer to fuse cues from both current and reference views.
Typically, these works flatten the feature maps into 1D token sequences, which are then fed into the transformer. 
In multi-view settings, tokens from all views are usually concatenated together to yield a long sequence. 
However, the dense global attention of transformers is computationally extensive. 
% with the computation complexity being quadratic to the length of input sequences. 
As a result, it is challenging to scale up these methods to high-resolution feature maps and many views. 
For example, the TransFusion \cite{ma2021transfusion} can only compute global attention between two views due to the large memory cost. 
Meanwhile, as empirically shown in Fig.\ref{fig:attn_map_demo}, the attention map of keypoints is very sparse, which only focuses on the body or the joint area. This is because the constraints among human keypoints tend to be adjacent and symmetric \cite{li2021tokenpose}. 
This observation also suggests that the dense attention among all locations in the image is relatively extravagant.

% P3: Our method
In this paper, we propose a compromised and yet efficient alternative to the global attention in pose estimation, named token-Pruned Pose Transformer (PPT). 
We calculate attention only within the human body area, rather than over the entire input image. 
Specifically, we select human body tokens and prune background tokens with the help of attention maps. As the human body only takes a small area of the entire image, the majority of input tokens can be pruned. 
We reveal that pruning these less informative tokens does not hurt the pose estimation accuracy, but can accelerate the entire networks. 
Interestingly, as a by-product, PPT can also predict a rough human mask without the guidance of ground truth mask annotations.

% Multi-view
Moreover, we extend PPT to multi-view settings. 
As in Fig.\ref{fig:cross_fusion_type}, previous cross-view fusion methods consider all pixels in the reference view (global fusion) or pixels along the epipolar line (epipolar-based fusion) as candidates. 
The former is computationally extensive and inevitably introduces noise from the background, and the latter requires accurate calibration and lacks semantic information. 
Built upon PPT, we propose a new fusion strategy, called \textit{human area fusion}, which considers  human foreground pixels as corresponding candidates.
% It is an efficient way to perform the cross-view fusion. Moreover, it also avoids the useless or even disturbing information from the background tokens and makes the model focus on the constraints within the human body. 
Specifically, we firstly use PPT to locate the human body tokens on each view, and then perform the multi-view fusion among these selected tokens with transformers. Thus, our method is an efficient fusion strategy and can easily be extended to many views.

Our main contributions are summarized as follows:
\begin{enumerate}
    
    \item We propose the token-Pruned Pose Transformer (PPT) for efficient 2D human pose estimation, which can locate the human body area and prune background tokens with the help of a Human Token Identification module. 
    
    \item We propose the strategy of ``Human area fusion" for multi-view pose estimation. Built upon PPT, the multi-view PPT can efficiently fuse cues from human areas of multiple views. 
    
    \item Experimental results on COCO and MPII demonstrate that our PPT can maintain the pose estimation accuracy while significantly reduce the computational cost. Results on Human 3.6M and Ski-Pose show that human area fusion outperforms previous fusion methods on 2D and 3D metrics. 
    
\end{enumerate}

\vspace{-0.5em}
\section{Related Work}
\vspace{-0.5em}
\subsection{Efficient Vision Transformers}

Recently, the transformer \cite{vaswani2017attention} achieves great progresses on many computer vision tasks, such as classification \cite{dosovitskiy2020image,touvron2020training}, object detection \cite{carion2020end,zhu2020deformable,fang2021you}, and semantic segmentation \cite{zheng2020rethinking,wang2020end,yan2022after,you2022class}. 
While being promising in accuracy, the vanilla ViT \cite{dosovitskiy2020image} is cumbersome and computationally intensive. Therefore, many algorithms have been proposed to improve the efficiency of vision transformers. 
Recent works demonstrate that some popular model compression methods such as network pruning \cite{han2015deep,chen2021chasing,chen2022principle,yu2022unified}, knowledge distillation \cite{hinton2015distilling,touvron2020training,chen2022dearkd}, and quantization \cite{shen2020q,sun2022vaqf} can be applied to ViTs. 
Besides, other methods introduce CNN properties such as hierarchy and locality into the transformers to alleviate the burden of computing global attention \cite{liu2021swin,chen2021crossvit}. 
On the other hand, some works accelerate the model by slimming the input tokens \cite{yuan2021tokens,caron2021emerging,ryoo2021tokenlearner,rao2021dynamicvit,kong2021spvit,liang2022evit,meng2022adavit}. 
Specifically, the Token-to-tokens \cite{yuan2021tokens} aims to reduce the number of tokens by aggregating neighboring tokens into one token. 
The TokenLearner \cite{ryoo2021tokenlearner} mines important tokens by learnable attention weights conditioned on the input feature. 
The DynamicViT \cite{rao2021dynamicvit} prunes less informative tokens with an extra learned token selector.
The EViT \cite{liang2022evit} reduces and reorganizes image tokens based on the classification token. 
However, all these models have only been designed for classification, where the final prediction only depends on the special classification token. 
% While in pose estimation, the spatial clues is rather important. 

\vspace{-0.5em}
\subsection{Human Pose Estimation}

\subsubsection{Monocular 2D Pose Estimation} 
In the past few years, many successful CNNs are proposed in 2D human pose estimation. 
They usually capture both low-level and high-level representations \cite{wei2016convolutional,chen2018cascaded,newell2016stacked,chu2017multi,xiao2018simple,sun2019deep}, or use the structural of skeletons to capture the spatial constraints \cite{tompson2014joint,ke2018multi,papandreou2018personlab,kong2019adaptive,kong2020rotation,chen2020nonparametric,kong2020sia}.
% However, the locality nature of convolution still makes it difficult to capture and model long-range relationships, which is nontrivial in pose estimation.  
% 
Recently, many works introduce transformers into pose estimation tasks \cite{yang2020transpose,li2021tokenpose,mao2021tfpose,li2021pose,lin2020end,zheng20213d}. 
Specifically, TransPose \cite{yang2020transpose} utilizes transformers to explain dependencies of keypoint predictions. TokenPose \cite{li2021tokenpose} applies additional keypoint tokens to learn constraint relationships and appearance cues. Both works demonstrate the necessity of global attention in pose estimation. 
\vspace{-0.5em}
\subsubsection{Efficient 2D Pose Estimation} 
Some recent works also explore efficient architecture design for real-time pose estimation \cite{osokin2018real,neff2020efficienthrnet,shen2021towards,wang2022lite,zhang2021efficientpose,yu2021lite}. 
For example, EfficientPose \cite{zhang2021efficientpose} designs an efficient backbone with neural architecture search. 
Lite-HRNet \cite{yu2021lite} proposes the conditional channel weighting unit to replace the heavy shuffle blocks of HRNet. 
However, these works all focus on CNN-based networks, and none of them study transformer-based networks. 
\vspace{-0.5em}
\subsubsection{Multi-view Pose Estimation}  
3D pose estimation from multiple views usually takes two steps: predicting 2D joints on each view separately with a 2D pose detector, and lifting 2D joints to 3D space via triangulation. 
Recently, many methods focus on enabling the 2D pose detector to fuse information from other views \cite{qiu2019cross,zhang2021adafuse,xie2020metafuse,he2020epipolar,ma2021transfusion}. 
They can be categorized into two groups: 
1) Epipolar-based fusion. The features of one pixel in one view is augmented by fusing features along the corresponding epipolar line of other views. 
Specifically, the AdaFuse \cite{zhang2021adafuse} adds the largest response on the heatmap along the epipolar line. 
The epipolar transformer \cite{he2020epipolar} applies the non-local module \cite{wang2018non} on intermediate features to obtain the fusion weights. 
However, this fusion strategy requires precise camera calibration and discard information outside the epipolar lines.
2) Global fusion. The features of one pixel in one view are augmented by fusing features of all locations in other views. 
In detail, the Cross-view Fusion \cite{qiu2019cross} learns a fixed attention matrix to fuse heatmaps in all other views. 
The TransFusion \cite{ma2021transfusion} applies the transformers to fuse features of the reference views and demonstrates that global attention is necessary. 
However, the computation complexity of global fusion is quadratic to the resolution of input images and number of views. 
Thus, both categories have their limitations. A fusion algorithm that can overcome these drawbacks and maintains their advantages is in need.

\vspace{-0.5em}
\section{Methodology}
\vspace{-0.5em}

\subsection{Token-Pruned Pose Transformer}
\label{sec:PPT_method}

\begin{figure}[!t]
    \centering
    \includegraphics[width=0.92\linewidth]{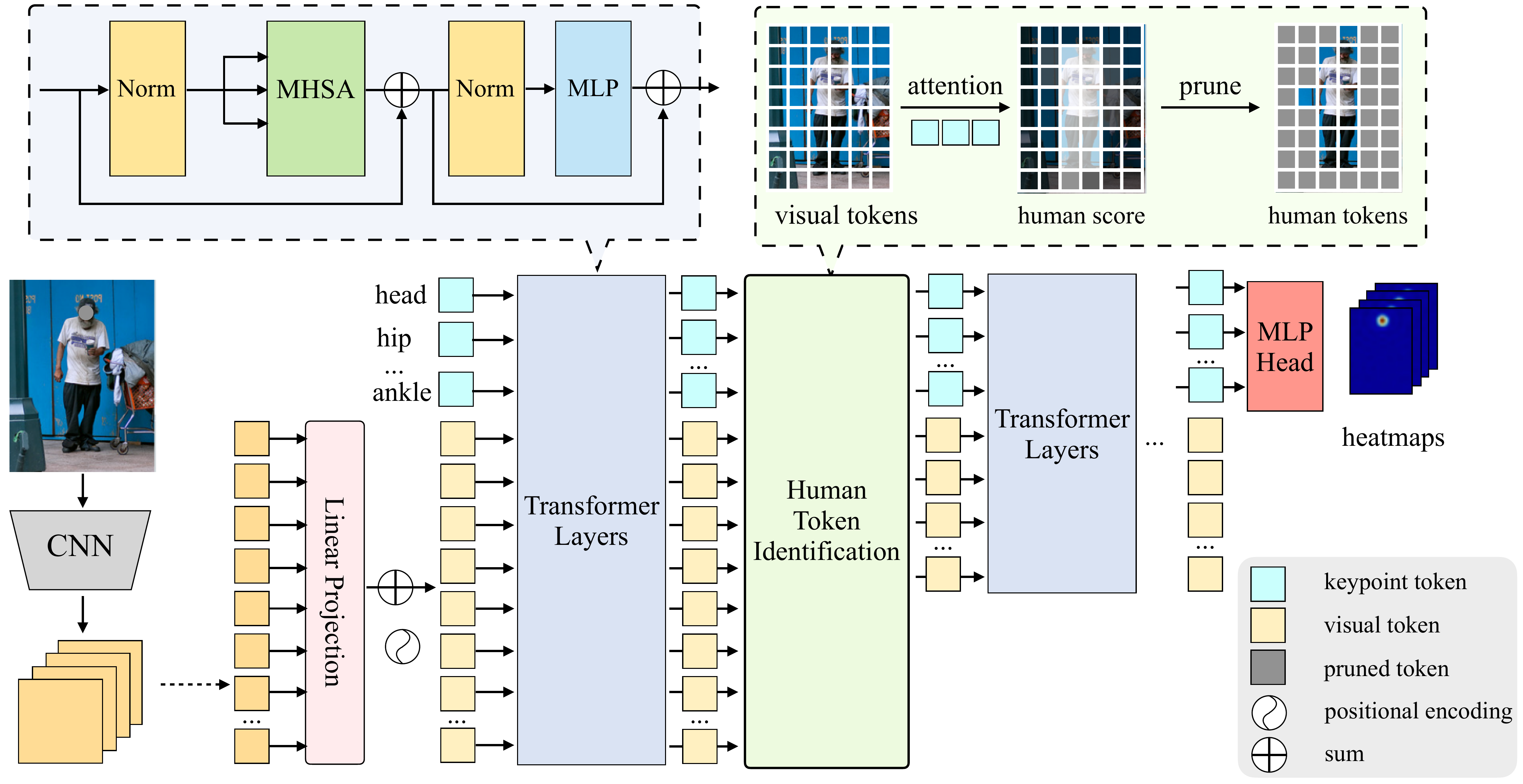}
    \caption{\small{Framework of the token-Pruned Pose Transformer (PPT). 
    The visual tokens are obtained from the flattened CNN feature maps. The keypoint tokens are added to represent each joint and predict the keypoints heatmaps. The Human Token Identification (HTI) module is inserted inside the transformer layers to locate human visual tokens and prune background tokens. Thus the followed transformer layers are only performed on these selected tokens. }  }
    \label{fig:2D_prune_framework}
\end{figure}

\subsubsection{Overview}
Fig.\ref{fig:2D_prune_framework} is an overview of our token-Pruned Pose Transformer. 
Following \cite{li2021tokenpose}, the input RGB image $\mathbf{I}$ first go through a shallow CNN backbone $\mathcal{B}(\cdot)$ to obtain the feature map $\mathbf{F} \in  \mathbb{R}^{ H \times W \times C}$.
Then $\mathbf{F}$ is decomposed into flattened image patches $\mathbf{F}_p \in \mathbb{R}^{N_v \times (C \cdot P_h \cdot P_w) }$, where $(P_h, P_w)$ is the resolution of each image patch, and $ N_v = \frac{H}{P_h} \cdot \frac{W}{P_w}$ is the total number of patches \cite{dosovitskiy2020image}. Then a linear projection is applied to project $\mathbf{F}_p$ into $\mathbf{X}_p \in \mathbb{R}^{N_v \times D}$, where $D$ is the dimension of hidden embeddings.
The 2D positional encodings $\mathbf{E} \in \mathbb{R}^{N_v \times D}$ are added to make the transformer aware of position information \cite{vaswani2017attention}, \ie  $\mathbf{X}_v = \mathbf{X}_p + \mathbf{E}$, namely the visual token.
Meanwhile, following TokenPose \cite{li2021tokenpose}, we have $J$ additional learnable keypoint tokens $\mathbf{X}_k \in \mathbb{R}^{J \times D}$ to represent $J$ target keypoints. 
The input sequence to the transformer is $\mathbf{X}^0 = [\mathbf{X}_k, \mathbf{X}_v] \in  \mathbb{R}^{N \times D}$, where $N = N_v + J$ and $[\ldots]$ is the concatenation operation.   

The transformer has $L$ encoder layers in total. 
At the $L_1^{th}$ layer, the Human Token Identification (HTI) module locates $K$ most informative visual tokens where human body appears and prunes the remaining tokens. 
We denote $r=\frac{K}{N_v} (0<r<1)$ as the keep ratio.
As a result, the length of the sequence is reduced to $N'=rN_v+J$ for the following transformer layers.  
The HTI is conducted $e$ times at the $L_1^{th}, L_2^{th}, \ldots, L_e^{th}$ layers. 
Thus, PPT can progressively reduce the length of visual tokens. 
Finally, the total number of tokens is $r^e N_v + J$. 
The prediction head projects the keypoint tokens in the last layer $\mathbf{X}_k^L \in \mathbb{R}^{J\times D}$ into the output heatmaps $\mathbf{H} \in \mathbb{R}^{J\times (H_h\cdot W_h)}$. 
\subsubsection{Transformer Encoder Layer. }
The encoder layer consists of the multi-headed self-attention (MHSA) and multi-layer perceptron (MLP). Operations in one encoder layer is shown in Fig. \ref{fig:2D_prune_framework}.
% 
% \begin{equation}
% \begin{aligned}
%     \mathbf{\hat{X}}^l &= \text{MHSA} ( \text{LN} ( \mathbf{X}^{l-1} ) ) +  \mathbf{X}^{l-1}, & & l=1 \ldots L \\
%     \mathbf{X}^l &= \text{MLP} ( \text{LN} (\mathbf{\hat{X}}^l)) + \mathbf{\hat{X}}^l, & & l=1 \ldots L
% \end{aligned}
% \end{equation}
% 
The self-attention  aims to match a query and a set of key-value pairs to an output \cite{vaswani2017attention}. 
Given the input $\mathbf{X}$, three linear projections are applied to transfer $\mathbf{X}$ into three matrices of equal size, namely the query $\mathbf{Q}$, the key $\mathbf{K}$, and the value $\mathbf{V}$.  
 The self-attention (SA) operation is calculated by:
\begin{equation}
  \text{SA}(\mathbf{X}) =  \text{Softmax}( \frac{ \mathbf{Q} \mathbf{K}^T }{\sqrt{D}}  )\mathbf{V},
  \label{eq:attention}
\end{equation}
For MHSA, $H$ self-attention modules are applied to $\mathbf{X}$ separately, and each of them produces an output sequence.

%  each keypoint token is a linear combination of all value vectors of visual token, where the combination coefficients is the attention values from the query vector for that keypoint token with respect to all visual tokens [24]. The attention value determines how much information of each visual token is fused into the output [41]. Thus, it is natural to assume that the attention value indicates the importance of each visual token, where tokens with joints should have large attention value. We will give more illustration in the final version. 

\subsubsection{Human Token Identification (HTI). }
The TokenPose \cite{li2021tokenpose} conducts self-attention among all visual tokens, which is cumbersome and inefficient. 
From Equation \ref{eq:attention}, we know that each keypoint token $\mathbf{X}_k^j$ interacts with all visual tokens $\mathbf{X}_v$ via the attention mechanism: 
\begin{equation}
    \text{Softmax} (\frac{ \mathbf{q}_k^j \mathbf{K}_v^T }{\sqrt{D}}) \mathbf{V}_v = \mathbf{a}^j \mathbf{V}_v,
\end{equation}
where $ \mathbf{q}_k^j$ denotes the query vector of $\mathbf{X}_k^j$, $\mathbf{K}_v$ and $\mathbf{V}_v$ are the keys and values of visual tokens $\mathbf{X}_v$. 
To this end, each keypoint token is a linear combination of all value vectors of visual tokens. The combination coefficients $\mathbf{a}^j \in \mathbb{R}^{N_v}$ are the attention values from the query vector for that keypoint token with respect to all visual tokens. 
To put it differently, the attention value determines how much information of each visual token is fused into the output. 
Thus, it is natural to assume that the attention value $\mathbf{a}^j$ indicates the importance of each visual token in the keypoint prediction \cite{liang2022evit}. Typically, a large attention value suggests that the target joint is inside or nearby the corresponded visual token.

With this assumption, we propose the Human Token Identification module to select informative visual tokens with the help of attention scores of keypoint tokens. 
However, each keypoint token usually only attends to a few visual tokens around the target keypoint. And some keypoint tokens (such as the eye and the nose) may attend to close-by or even the same visual tokens. Thus, it is difficult to treat the attention values of each keypoint separately. 
For simplicity, as all human keypoints make up a rough human body area, we use $\mathbf{a} = \sum_j \mathbf{a}^j$ as the criterion to select visual tokens, which is the summation of all joints' attention maps. 
In detail, we keep visual tokens with the $K$ largest corresponding values in $\mathbf{a}$ as the human tokens, and prune the remaining tokens. As a result, only $K$ visual tokens and $J$ keypoint tokens are sent to the following layers.

\begin{figure}[t!]
    \centering
    \includegraphics[width=0.98\linewidth]{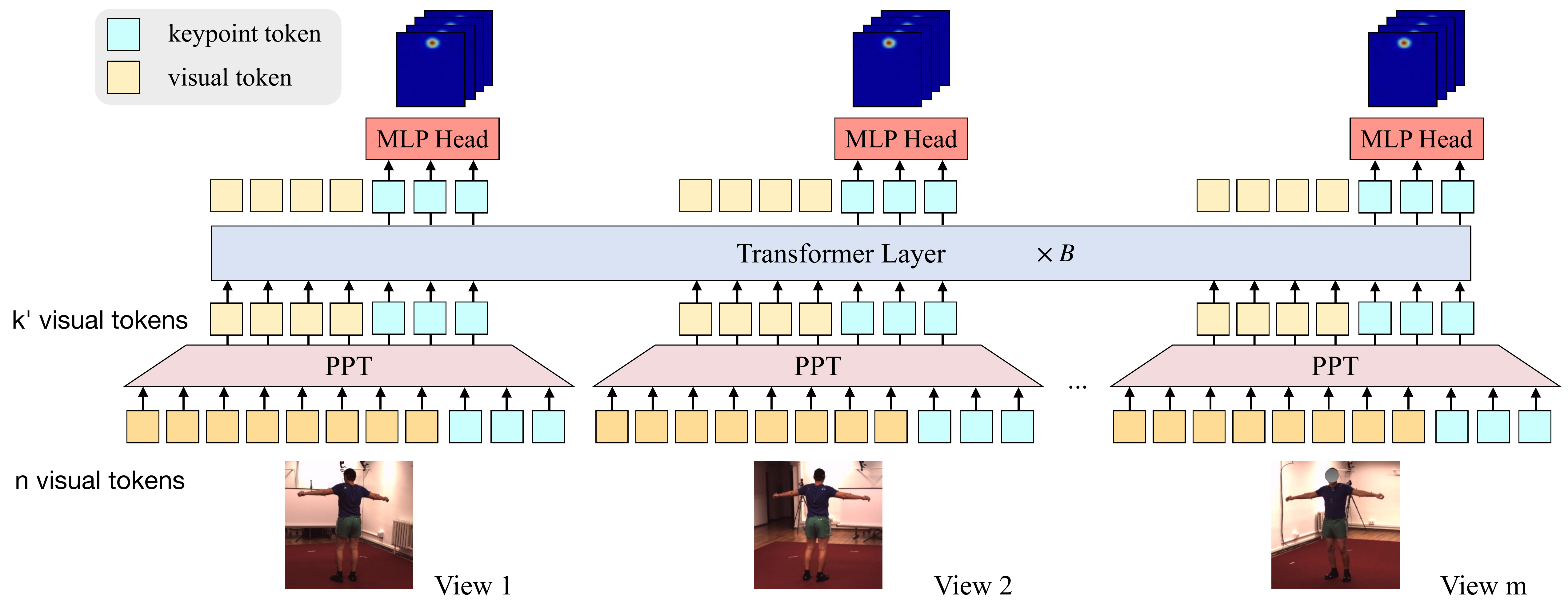}
    \caption{ \small{Overall framework of the Multi-view PPT. A share-weight PPT is applied to extract a subset of visual tokens for each view. Then $B$ transformer layers are applied to the concatenated tokens from each view to perform cross-view fusion. The output head takes keypoint tokens in each view to predict heatmaps.} }
    \label{fig:multi_view_framework}
\end{figure}

\vspace{-0.5em}
\subsection{Multi-view Pose Estimation with PPT}
\label{sec:PPT_multiview}
% \vspace{-0.3em}

\subsubsection{Human Area Fusion. }
We propose the concept of \textit{Human area fusion} for cross-view fusion in multi-view pose estimation, which considers pixels where human appears as corresponding candidates.
Suppose there are $m$ cameras, and each view maintains $n$ pixels (tokens) in its feature map. 
We summarize three typical types of cross-view fusion strategies in Fig.\ref{fig:cross_fusion_type}. 
1) For global fusion, each pixel in each view calculates attention with respect to all $n$ pixels in feature maps of other $m-1$ views. Thus the computational complexity is $\mathcal{O}(m^2n^2)$.
2) For epipolar-based fusion, each pixel in each view calculates attention with $k (k \ll n)$ pixels along the corresponded epipolar lines of other $m-1$ views. Thus the computational complexity is $\mathcal{O}(m^2nk)$. 
3) For our human area fusion, we firstly select $k'$ human foreground pixels in each view. Then we perform dense attention among these foreground tokens. As we also reduce the number of query pixels, the computational complexity is $\mathcal{O}(m^2k'^2)$. Typically, $k < k' \ll n$. 
Thus, our method is an efficient way to perform cross-view fusion. Moreover, it also avoids the useless or even disturbing information from the background tokens and thus makes the model focus on the constraints within the human body. 

% % 
% For each pixel in each view, the global fusion is calculated with respect to all $n$ pixels in feature maps of other $m-1$ views. Thus the computational complexity is $\mathcal{O}(m^2n^2)$. 
% % 
% Given an pixel in one view, methods with epipolar constraints \cite{he2020epipolar,zhang2021adafuse} only calculate attention with $k (k \ll n)$ pixels along the corresponded epipolar lines of other $m-1$ views. Thus the computational complexity is $\mathcal{O}(m^2nk)$. 
% % 
% % 
% For our human area fusion,  we firstly select a subset of $k'$ foreground pixels in each view. 
% For each foreground pixel in one view, we calculate attention with $k'$ pixels in other $m-1$ views. Thus, the computational complexity is $\mathcal{O}(m^2k'^2)$. Typically, $k < k' \ll n$. Thus, our method can accelerate the fusion while maintaining rich semantic information in each view.  

\subsubsection{Multi-view PPT. }
Naturally, we can apply an off-the-shelf segmentation network \cite{he2017mask} to obtain human foreground pixels and then perform human area fusion. However, a large amount of densely annotated images are required to train a segmentation model. 
To this end, we utilize PPT to efficiently locate a rough human foreground area without any mask labels, and further propose the \textit{multi-view PPT} for multi-view pose estimation. 
Specifically, we design our network in a two-stage paradigm, as shown in Fig.\ref{fig:multi_view_framework}.  
Given the image $\mathbf{I}^m$ in each view, the share-weight PPT firstly produces selected human tokens $\mathbf{ \tilde{X}}^m_v$ and keypoint tokens $\mathbf{X}^m_k$. 
Then we concatenate tokens from all views together and perform the dense attention among them with $B$ transformer encoder layers. 
To help the network perceive the 3D space information, we also add the 3D positional encodings \cite{ma2021transfusion} on all selected visual tokens. 
Thus, each keypoint token can fuse visual information from all views.  Moreover, it can learn correspondence constraints between keypoints both in the same view and among different views. 
Finally, a share-weight MLP head is placed on top of the keypoint token of each view to predicts keypoint heatmaps.

% tokens from all views are concatenated together as the input to the transformers.

% In the second stage, $B$ transformer layers are applied to perform the cross-view fusion.  
% In detail, tokens from all views are concatenated together as the input to the transformers.
% % \ie $\mathbf{X} = [\mathbf{X}^1_k, \mathbf{X}^2_k, \ldots, \mathbf{X}^M_k,  \mathbf{ \tilde{X}}^1_v, \mathbf{\tilde{X}}^2_v, \ldots \mathbf{ \tilde{X}}^M_v]$. 
% With self-attention calculated among these tokens, keypoint tokens in each view can fuse visual information from all views. Moreover, it can learn correspondence relations between keypoints both in the same view and among different views. 

\vspace{-0.5em}
\section{Experiments on monocular image}
\vspace{-0.5em}

\subsection{Settings} 

\subsubsection{Datasets \& Evaluation Metrics. } 
We firstly evaluate PPT on monocular 2D human pose estimation benchmarks. 
COCO \cite{lin2014microsoft} contains $200K$ images in the wild and $250K$ human instances with $17$ keypoints. Following top-down methods \cite{xiao2018simple,sun2019deep,li2021tokenpose}, we crop human instances with the ground truth bounding boxes for training and with the bounding boxes provided by SimpleBaseline \cite{xiao2018simple} for inference.   
The evaluation is based on object keypoint similarity, which measures the distance between the detected keypoint and the corresponding ground truth. The standard average precision (AP) and recall (AR) scores are reported.  
MPII \cite{andriluka14cvpr} contains about $25K$ images and $40K$ human instances with $16$ keypoints. The evaluation is based on the head-normalized probability of correct keypoint (PCKh) score \cite{andriluka14cvpr}. 
A keypoint is correct if it falls within a predefined threshold to the groundtruth location. We report the PCKh@0.5 score by convention.

\vspace{-0.5em}
\subsubsection{Implementation Details. } 
For fair comparison, we build our PPT based upon TokenPose-S, TokenPose-B, and TokenPose-L/D6 \cite{li2021tokenpose}, namely PPT-S, PPT-B, and PPT-L/D6, respectively. 
For PPT-S and PPT-B, the number of encoder layers $L$ is set to $12$, the embedding size $D$ is set to $192$, the number of heads $H$ is set to $8$. 
They take the shallow stem-net and the HRNet-W32 as the CNN backbone, respectively.  
Following \cite{rao2021dynamicvit,liang2022evit}, the HTI is performed $e=3$ times and is inserted before the $4^{th}$, $7^{th}$, and $10^{th}$ encoder layers. 
The PPT-L/D6 has $L=12$ encoder layers and takes HRNet-W48 as the backbone. the HTI is inserted before the $2^{th}$, $4^{th}$, and $5^{th}$ encoder layers. 
The number of visual tokens $N_v$ is $256$ for all networks, and the keep ratio $r$ is set to $0.7$ by default. Thus, only $88$ visual tokens are left after three rounds pruning.
We follow the same training recipes as \cite{li2021tokenpose}. 
In detail, all networks are optimized by Adam optimizer \cite{kingma2014adam} with Mean Square Error (MSE) loss for $300$ epochs. 
The learning rate is initialized with $0.001$ and decays at the $200$-th and the $260$-th epoch with ratio $0.1$. 
As locating human is difficult at early training stages, the keep ratio is gradually reduced from $1$ to $r$ with a cosine schedule during the early $100$ epochs.

\vspace{-0.5em}
\subsection{Results}
\vspace{-0.5em}

\begin{table}[!t]
% >>>>>>>>>>>>>>>>>>>>>>>>>>> COCO >>>>>>>>>>>>>>>>>>>>>>>>>>>
\centering
\resizebox{1.0\textwidth}{!}{
\begin{tabular}{l|l|l|l|lccccc}
\toprule
Method &  \#Params & GFLOPs  & GFLOPs$^T$ & AP & AP$^{50}$ & AP$^{75}$ & AP$^{M}$ & AP$^{L}$ & AR \\
\hline
SimpleBaseline-Res50 \cite{xiao2018simple}   & 34M &   8.9  & - &  70.4  &  88.6  &  78.3   &  67.1  &  77.2  & 76.3   \\
SimpleBaseline-Res101 \cite{xiao2018simple}  & 53M    & 12.4 & - & 71.4 & 89.3 & 79.3 & 68.1 & 78.1 & 77.1   \\
SimpleBaseline-Res152  \cite{xiao2018simple} & 68.6M  & 15.7 & - & 72.0 & 89.3 & 79.8 & 68.7 & 78.9 & 77.8   \\
HRNet-W32 \cite{sun2019deep}                & 28.5M  & 7.1  & - & 74.4 & 90.5 & 81.9 & 70.8  & 81.0 & 79.8  \\
HRNet-W48 \cite{sun2019deep}                & 63.6M  & 14.6 & - & 75.1 & 90.6 & 82.2 & 71.5 & 81.8 & 80.4  \\

\hline 
Lite-HRNet-18 \cite{yu2021lite} & 1.1M & 0.20 & - & 64.8 & 86.7 & 73.0 & 62.1 & 70.5 & 71.2 \\
Lite-HRNet-30 \cite{yu2021lite} & 1.8M & 0.31 & - & 67.2 & 88.0 & 75.0 & 64.3 & 73.1 & 73.3 \\
EfficientPose-B \cite{zhang2021efficientpose} & 3.3M & 1.1 & - & 71.1  & - & - & - & - & -  \\
EfficientPose-C \cite{zhang2021efficientpose} &  5.0M & 1.6 & - & 71.3 & - & - & - & - & -  \\

\hline 
TransPose-R-A4 \cite{yang2020transpose} & 6.0M & 8.9 & 3.38 & 72.6 & 89.1 & 79.9 & 68.8 & 79.8 & 78.0 \\
TransPose-H-S \cite{yang2020transpose} & 8.0M & 10.2 & 4.88 & 74.2 & 89.6 & 80.8 & 70.6 & 81.0 & 79.5 \\
TransPose-H-A6 \cite{yang2020transpose} & 17.5M & 21.8 & 11.4 & 75.8 & 90.1 & 82.1 & 71.9 & 82.8 & 80.8 \\
TokenPose-S \cite{li2021tokenpose}   & 6.6M & 2.2 &         1.44    &        72.5 & 89.3 & 79.7 & 68.8 & 79.6 & 78.0 \\
TokenPose-B \cite{li2021tokenpose}   & 13.5M &  5.7 & 1.44 & 74.7 & 89.8 & 81.4 & 71.3 & 81.4 & 80.0 \\
TokenPose-L/D6 \cite{li2021tokenpose}  & 20.8M & 9.1 & 0.72 & 75.4 & 90.0 & 81.8 & 71.8 & 82.4 & 80.4 \\
\hline

PPT-S (ours)                         & 6.6M & 1.6(\textbf{-27\%}) &  0.89(-38\%) & 72.2(\textbf{-0.3}) & 89.0 & 79.7 & 68.6 & 79.3 & 77.8 \\
PPT-B  (ours)                        & 13.5M &  5.0(\textbf{-12\%}) & 0.89(-38\%) & 74.4(\textbf{-0.3}) & 89.6 & 80.9 & 70.8 & 81.4 & 79.6 \\
PPT-L/D6 (ours) & 20.8M & 8.7(\textbf{-4\%}) & 0.50(-31\%) & 75.2(\textbf{-0.2}) & 89.8 & 81.7 & 71.7 & 82.1 & 80.4 \\
\bottomrule
\end{tabular}
}
\caption{ \small{Results on COCO validation dataset. The input size is $256\times192$. GFLOPs$^T$ means the GFLOPs for the transformers only following equations from \cite{kong2021spvit}, as our method only focus on accelerating the transformers. } }
\label{tab:coco_val}
\end{table}

\begin{table}[!t]
% >>>>>>>>>>>>>>>>>>>>>>>>>>> MPII >>>>>>>>>>>>>>>>>>>>>>>>>>>
\centering
\resizebox{1.0\textwidth}{!}{
\begin{tabular}{l|l|l|lllllll|c}
\toprule
Method & \#Params & GFLOPs & Head & Sho & Elb & Wri & Hip & Kne & Ank & Mean  \\
\hline
SimpleBaseline-Res50 \cite{xiao2018simple}  & 34M & 12.0 & 96.4 & 95.3 & 89.0 & 83.2 & 88.4 & 84.0 & 79.6 & 88.5  \\
SimpleBaseline-Res101 \cite{xiao2018simple} & 53M & 16.5 & 96.9 & 95.9 & 89.5 & 84.4 & 88.4 & 84.5 & 80.7 & 89.1 \\
SimpleBaseline-Res152 \cite{xiao2018simple} & 53M & 21.0 & 97.0 & 95.9 & 90.0 & 85.0 & 89.2 & 85.3 & 81.3 & 89.6 \\
HRNet-W32. \cite{sun2019deep}               & 28.5M & 9.5 & 96.9 & 96.0 & 90.6 & 85.8 & 88.7 & 86.6 & 82.6 & 90.1 \\
\hline
TokenPose-S \cite{li2021tokenpose}   & 7.7M  & 2.5 & 96.0 & 94.5 & 86.5 & 79.7 & 86.7 & 80.1 & 75.2 & 86.2 \\
PPT-S                                & 7.7M  & 1.9 (-\textbf{24\%})  & 96.6 & 94.9 & 87.6 & 81.3 & 87.1 & 82.4 & 76.7 & 87.3 (+\textbf{1.1}) \\
TokenPose-B \cite{li2021tokenpose}  & 14.4M &  7.1 & 97.0  & 96.1   &  90.1  & 85.6   & 89.2   & 86.1  & 80.3 & 89.7  \\
PPT-B                               & 14.4M  & 6.2 (-\textbf{13\%}) & 97.0 & 95.7 & 90.1 & 85.7 & 89.4 &  85.8 & 81.2 & 89.8 (+\textbf{0.1})\\
\bottomrule
\end{tabular}
}
\caption{ \small{Results on the MPII validation set (PCKh@0.5).  The input size is $256\times256$.}}
\label{tab:mpii_2d}
\end{table}

The results are shown in Table \ref{tab:coco_val} and Table \ref{tab:mpii_2d} for COCO and MPII, respectively. 
Generally, the transformer-based methods \cite{li2021tokenpose,yang2020transpose} maintain less number of parameters. 
% >>>>>>>>>>>>>>>>>>>>> COCO >>>>>>>>>>>>>>>>>>>>>
On COCO, compared with the TokenPose, PPT achieves significant acceleration while matching its accuracy. 
For example, PPT-S reduces 27\% total inference FLOPs while only reducing $0.3$ AP. Compared to SimpleBaseline-ResNet152 \cite{xiao2018simple}, PPT-S achieves equal performance but only requires $10\%$ FLOPS. 
We can also observe consistent conclusion on PPT-B and PPT-L. 
Note that, for PPT-B and PPT-L, the CNN backbone takes a large portion of computation. Thus, the reduction of total FLOPs is relatively small. 
% >>> Efficient Pose Estimation Network >>>
Meanwhile, compared with other efficient pose estimation networks \cite{yu2021lite,zhang2021efficientpose}, the AP of PPT-S is $72.2$, which is much better than EfficientPose-C \cite{zhang2021efficientpose}  with $71.3$ AP at the same FLOPs level. 
% >>>>>>>>>>>>>>>>>>>>> MPII >>>>>>>>>>>>>>>>>>>>>
More over, On MPII, our PPT-S can even improve on the PCKh of TokenPose-S by 1.1\%. 
We believe that slimming the number of tokens can also make the attention focus on key elements \cite{zhu2020deformable}. 
Thus, our PPT is efficient yet powerful, and it is applicable to any TokenPose variants. 
All of these results suggest that pruning background tokens does not hurt the overall accuracy and calculating attention among human foreground tokens is sufficient for 2D human pose estimation.

\vspace{-0.5em}
\subsection{Visualizations}
\vspace{-0.5em}

\begin{figure}[t]
% >>>>>>>>>>>>>>>>>>>>>>>>>>>>>>>>>>>>>>>>>>>>>>>>>>>>>>>>>>>>>>>>>>>>>>>>>>>>>>>>>>
    \small 
    \centering
    \subfigure[]{
    \label{Fig.mask.0}
    \includegraphics[width=0.45\textwidth]{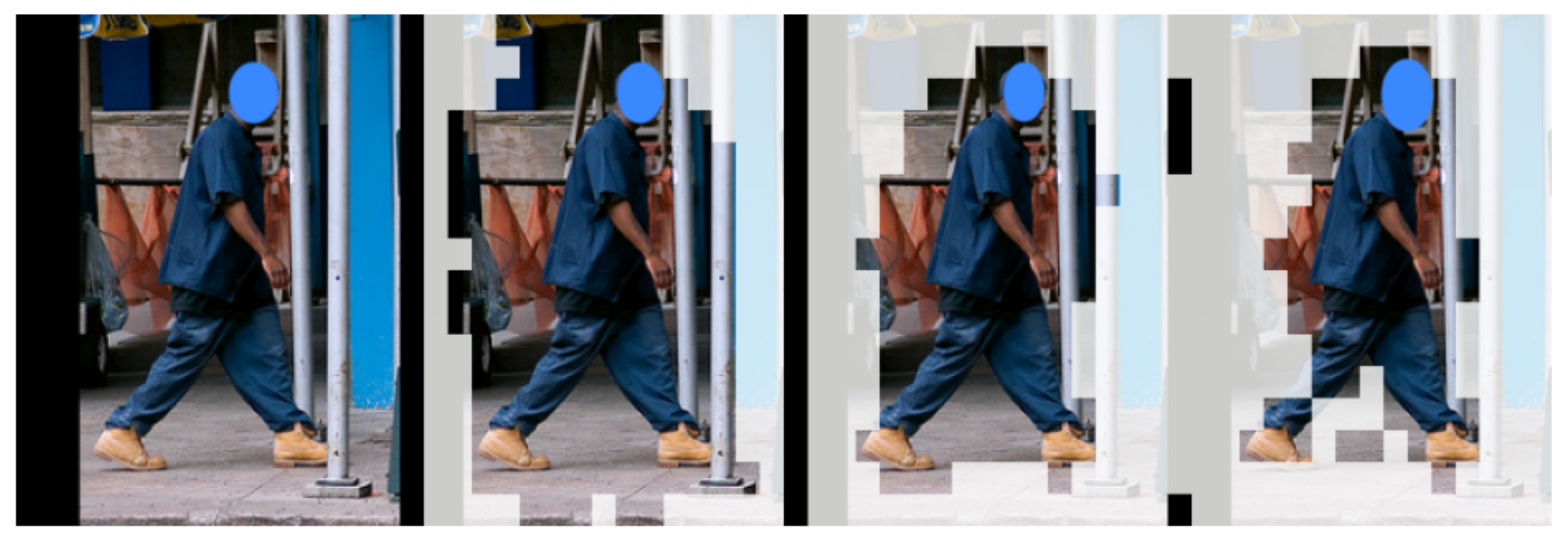}}  % label 0
    \subfigure[]{
    \label{Fig.mask.1}
    \includegraphics[width=0.45\textwidth]{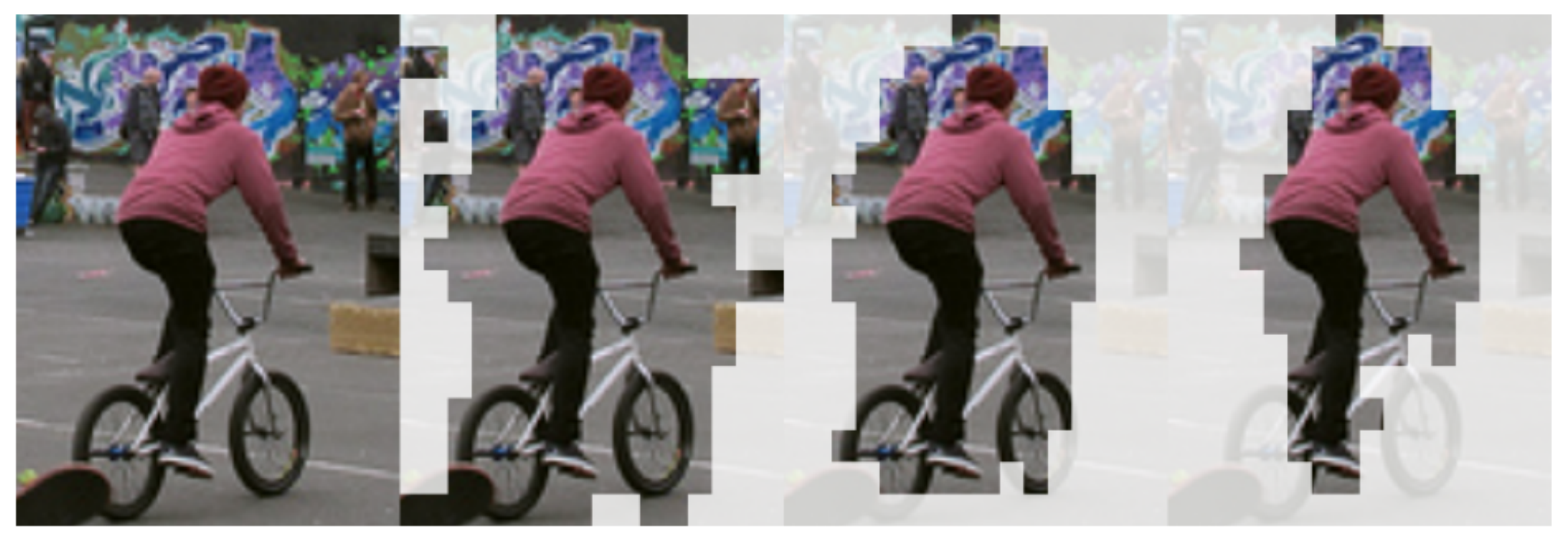}} % label 1
    \subfigure[]{
    \label{Fig.mask.2}
    \includegraphics[width=0.45\textwidth]{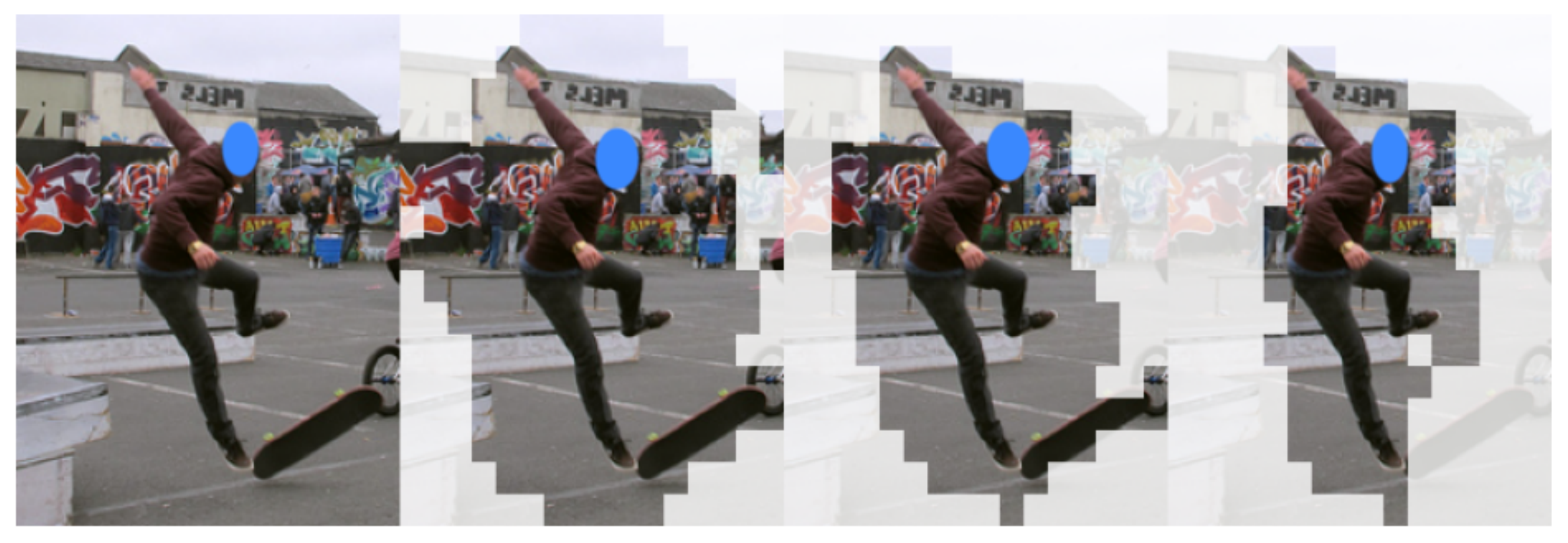}} % label 2
    \subfigure[]{
    \label{Fig.mask.3}
    \includegraphics[width=0.45\textwidth]{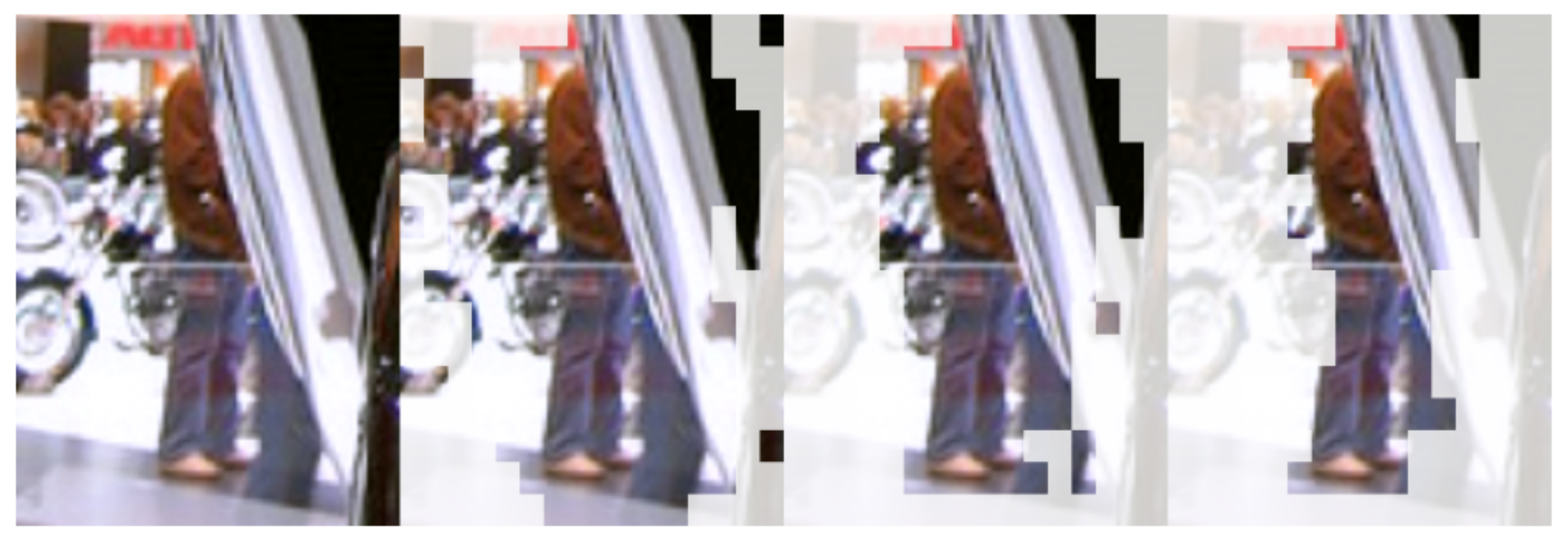}} % label 4
    \subfigure[]{
    \label{Fig.mask.4}
    \includegraphics[width=0.45\textwidth]{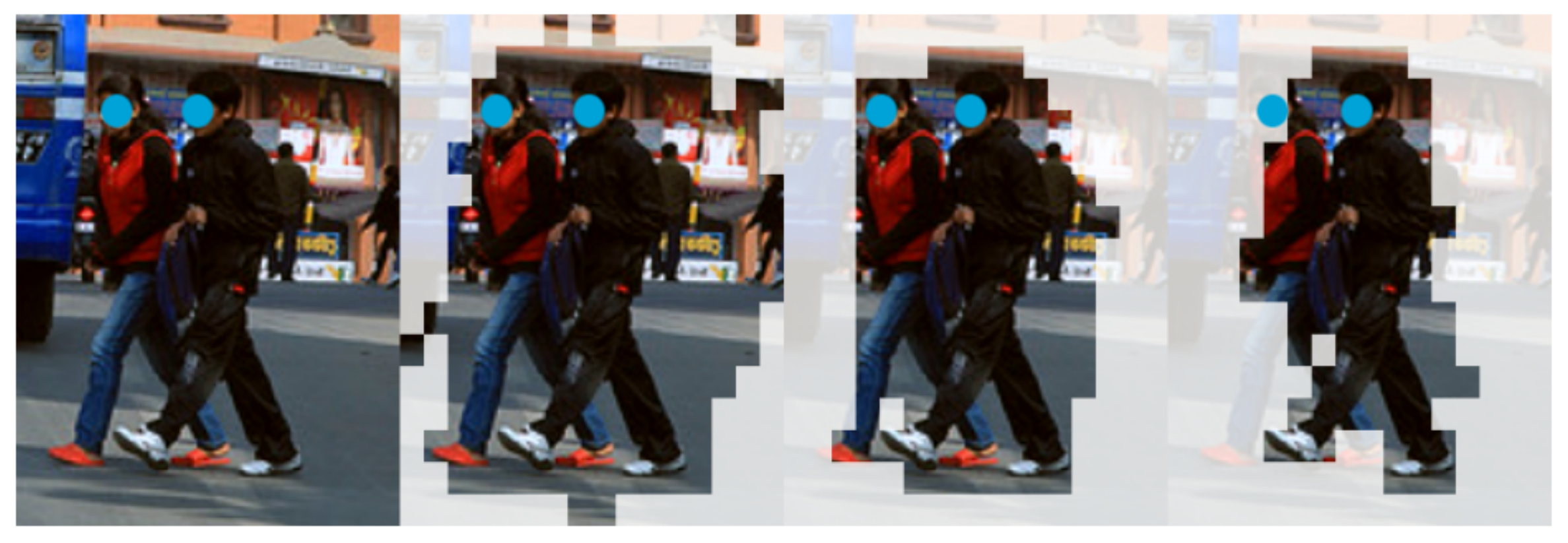}} % label 5    
    \subfigure[]{
    \label{Fig.mask.5}
    \includegraphics[width=0.45\textwidth]{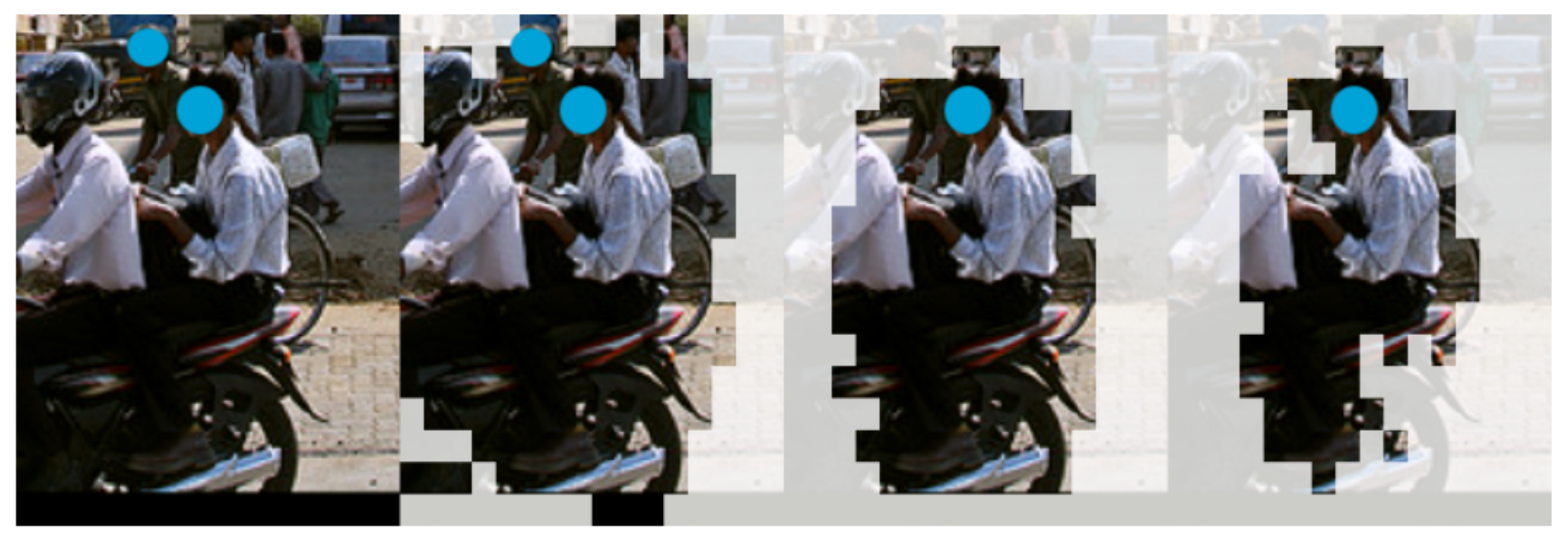}} % label 3
    \subfigure[]{
    \label{Fig.mask.6}
    \includegraphics[width=0.45\textwidth]{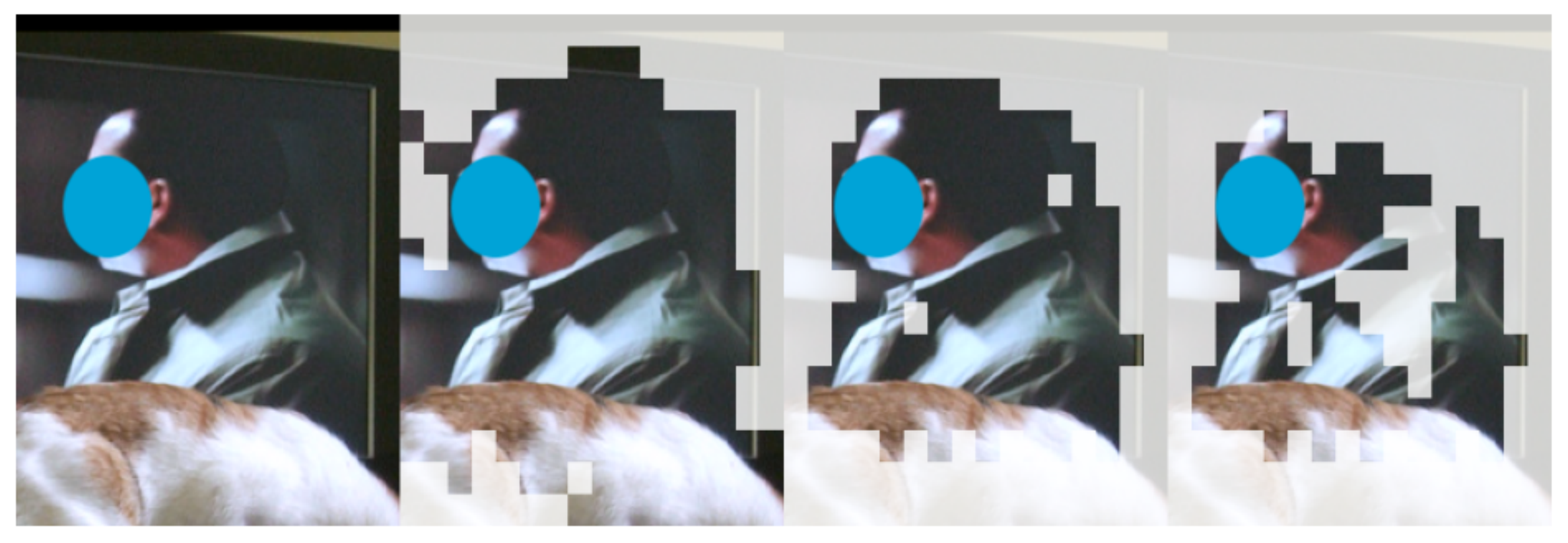}} % label 4
    \subfigure[]{
    \label{Fig.mask.7}
    \includegraphics[width=0.45\textwidth]{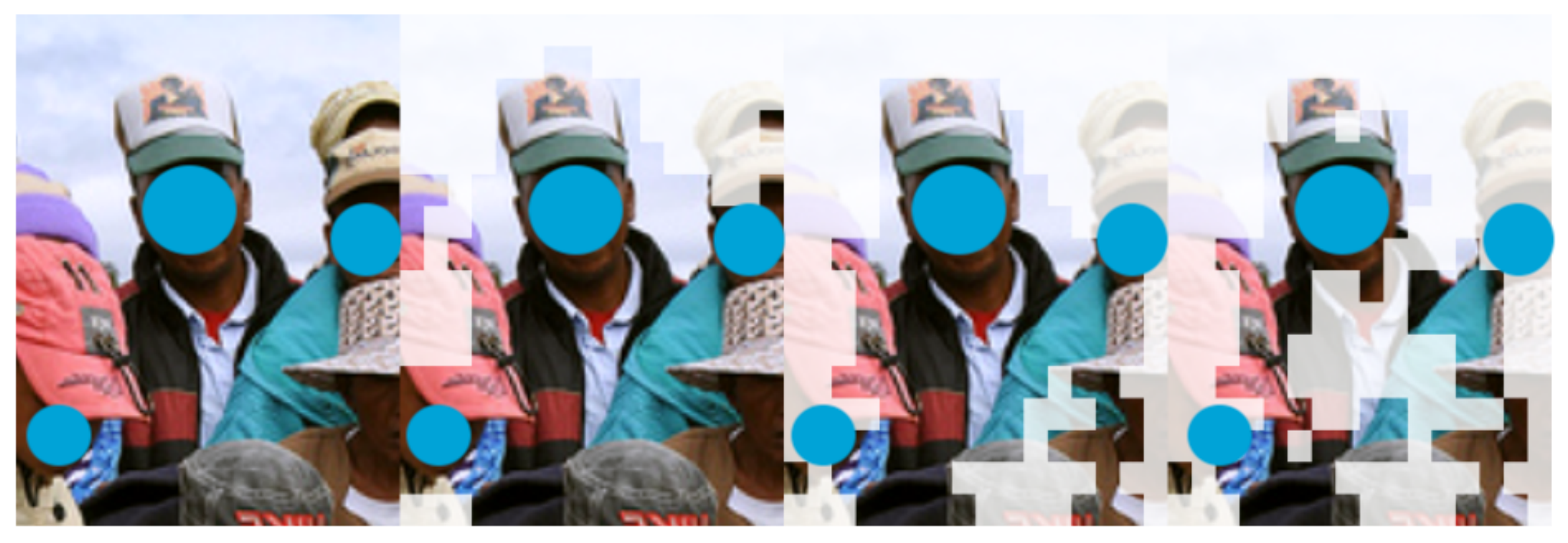}} % label 5
    \caption{
    \small{Visualizations of the selected tokens at each HTI module on COCO. The masked regions represent the pruned tokens (We use blue circles to mask out face for privacy issue). For each image group, the first column is the original image, the 2nd, 3rd, and 4th colums are the selected tokens by HTI at the $4^{th}$,$7^{th}$, and $10^{th}$ layers, respectively.  }
    }
    \label{fig:human_mask}
\end{figure}

We visualize the selected tokens from PPT-S in Fig. \ref{fig:human_mask}. We present the original images and the selected tokens at different layers.  
Remarkably, the human areas are gradually refined as the network deepens. The final selected tokens can be considered as a rough human mask. Thus, our HTI can successfully locate human tokens as expected. 
Moreover, the HTI can handle quite a few complicated situations such as man-object interaction (Fig.\ref{Fig.mask.1}), oblique body pose (Fig. \ref{Fig.mask.2}), occlusion (Fig. \ref{Fig.mask.3}), and multiple persons (Fig.\ref{Fig.mask.4} \ref{Fig.mask.5}). 
Nevertheless, when only part of human body appears in the image (Fig.\ref{Fig.mask.6}\ref{Fig.mask.7}), the quality of the located human mask could be imperfect. 
In these cases, we hypothesize that some keypoint tokens such as ankle and knee cannot locate the corresponding joints as they are invisible. Thus, they may just give equal attention score, which leads to inaccurate token selection.

\vspace{-0.5em}
\subsection{Ablation Studies}
\vspace{-0.5em}

The keep ratio $r$ controls the trade-off between the acceleration and the accuracy. Meanwhile, reducing tokens also introduces some regularization \cite{zhu2020deformable}. 
We take PPT-S and vary $r$ from $0.6$ to $0.8$ on both COCO and MPII. The results are shown in Table \ref{tab:diff_ratio}. The reduction of AP is always less than 1\%. 
When the $r$ is relatively small, PPT can achieve considerable speedup but may not cover the entire human body. As a result, the accuracy of pose estimation is slightly dropped.  
To maintain the accuracy, we choose $0.7$ as our default keep ratio.

\begin{table}[ht]
\centering
\resizebox{0.99\textwidth}{!}{
\begin{tabular}{c|c|c|ccc|ccc}
\toprule
\multirow{2}{*}{Method}  & \multirow{2}{*}{Keep Ratio} & \multirow{2}{*}{\# Visual Tokens} & \multicolumn{3}{c|}{COCO}       & \multicolumn{3}{c}{MPII}             \\ 
\cline{4-9}
& & & AP & AR & FLOPs & PCKh@0.5 & PCKh@0.1 & FLOPs \\ 
\hline
TokenPose-S & 1.0     & 256 (100\%) & 72.5   &  78.0 &  2.23   &     86.2  & 32.2   &  2.53     \\ 
PPT-S & 0.8 & 131 (51\%) & 72.0 (-0.5)   &  77.6(-0.4)    &  1.75 (-22\%) &  86.9 (+0.7)   &  32.9 (+0.7) &  2.06 (-19\%)      \\ 
PPT-S & 0.7 & 88 (34\%)  & 72.2 (-0.3) & 77.8 (-0.2) &  1.61 (-27\%)     &  87.3 (+1.1)  & 34.1 (+1.9)  & 1.92 (-24\%) \\
PPT-S & 0.6 & 56 (22\%)  & 71.8 (-0.7)  &  77.5 (-0.5) &  1.52 (-32\%)    &  86.7 (+0.5)  & 32.3 (+0.1) &   1.82 (-28\%)     \\
\bottomrule
\end{tabular}
}
\caption{ \small{Results of PPT-S on COCO and MPII with different keep ratio $r$.} }
\label{tab:diff_ratio}
\end{table}

\vspace{-0.5em}
\section{Experiments on Multi-view Pose Estimation }
\vspace{-0.5em}
\subsection{Settings}
\vspace{-0.5em}
\subsubsection{Datasets \& Evaluation Metrics.} 
We evaluate multi-view PPT on two single-person datasets of multi-view 3D human pose estimation, \ie Human 3.6M \cite{h36m_pami,IonescuSminchisescu11} and Ski-Pose \cite{sporri2016reasearch,fasel2017joint} \footnote{Only authors from UCI downloaded and accessed these two datasets. Authors from Tencent and Meta don't have access to them. }. 
Human 3.6M contains video frames captured by $M=4$ indoor cameras. 
It includes many daily activities such as eating and discussion. We follow the same train-test split as in \cite{qiu2019cross,iskakov2019learnable,he2020epipolar}, where subjects $1, 5, 6, 7, 8$ are used for training, and $9, 11$ are for testing. 
We also exclude some scenes of  $S9$ from the evaluation as their 3D annotations are damaged \cite{iskakov2019learnable}. 
Ski-Pose contains video frames captured by outdoor cameras. It is created to help analyze skiers's giant slalom. There are $8,481$ and $1,716$ frames in the training and testing sets, respectively. 
We use the Joint Detection Rate (JDR) on original images \cite{qiu2019cross} to evaluate the 2D pose accuracy. 
JDR measures the percentage of successfully detected keypoints within a predefined distance of the ground truth location. 
The 3D pose is evaluated by Mean Per Joint Position Error (MPJPE) between the ground truth 3D pose in world coordinates and the estimated 3D pose.

\subsubsection{Implementation Details. } 
We build multi-view PPT upon PPT-S. The first $9$ transformer layers are used to extract human tokens, and the last $3$ transformer layers are used for cross-view fusion. Thus, no additional parameters are introduced. 
Following the settings in \cite{he2020epipolar,ma2021transfusion}, we start from a PPT-S pre-trained on COCO and finetune it on multi-view human pose datasets, as it is difficult to train the transformer from scratch with examples in limited scenes. 
We apply Adam optimizer and train the model for $20$ epochs with MSE loss. The learning rate starts with $0.001$ and later on decays at $10$-th and $15$-th epoch with ratio $0.1$. The keep ratio $r$ is set to $0.7$ through the entire training process. 
We resize input images to $256\times256$ and follow the same data augmentation in \cite{qiu2019cross,ma2021transfusion}.

\begin{table}[t]
% >>>>>>>>>>>>>>>>>>>>>>>>>>>>>>>>>>> Human 3.6M 2D >>>>>>>>>>>>>>>>>>>>>>>>>>>>>>>>>>>
\centering
\resizebox{1.0\textwidth}{!}{
\begin{tabular}{l|c|l|ccccccccccc|c}
\toprule
Method              & \#V   & MACs  & shlder & elb & wri & hip & knee & ankle & root & belly & neck & nose & head & Avg  \\
\midrule
ResNet50 \cite{xiao2018simple}   &   1    & 51.7G   &97.0 & 91.9 & 87.3 & 99.4 & 95.0 & 90.8 & 100.0 & 98.3 & 99.4 & 99.3 & 99.5  & 95.2 \\
TransPose \cite{yang2020transpose}  & 1   & 43.6G    & 96.0 & 92.9 & 88.4 & 99.0 & 95.0 & 91.8 & 100.0 & 97.5 & 99.0 & 99.4 & 99.6  & 95.3 \\
TokenPose \cite{li2021tokenpose} & 1 & 11.2G & 96.0 & 91.3 & 85.8 & 99.4 & 95.2 & 91.5 & 100.0 & 98.1 & 99.1 & 99.4 & 99.1 & 94.9 \\
\hline
Epipolar Transformer \cite{he2020epipolar}  & 2 & 51.7G &  97.0 & 93.1 & 91.8 & 99.1 & 96.5 & 91.9 & 100.0 & 99.3 & 99.8 & 99.8 & 99.3 & 96.3 \\
TransFusion \cite{ma2021transfusion}   & 2   & 50.2G       &  97.2 & 96.6 & 93.7 & 99.0 & 96.8 & 91.7 & 100.0 & 96.5 & 98.9 & 99.3 & 99.5 & 96.7   \\
Crossview Fusion \cite{qiu2019cross} & 4 & 55.1G   & 97.2 & 94.4 & 92.7 & \textbf{99.8} & 97.0 & 92.3 & 100.0 & 98.5 & 99.1 & 99.1 & 99.1  & 96.6 \\
\hline
TokenPose+Transformers & 4 & 11.5G & 97.1 & 97.3 & 95.2 & 99.2 & 98.1 & 93.1 & 100.0 & 98.8 & 99.2 & 99.3 & 99.1 & 97.4 \\

PPT & 1 & 9.6G & 96.0 & 91.8 & 86.5 & 99.2 & 95.6 & 92.2 & 100.0 & 98.4 & 99.3 & 99.5 & 99.4 & 95.3 \\
Multi-view PPT & 2 & 9.7G & 97.1 & 95.5 & 91.9 & 99.4 & 96.4 & 92.1 & 100.0 & 99.0 & 99.2 & 99.3 & 99.0 & 96.6  \\
Multi-view PPT & 4 & 9.7G & 97.6 & \textbf{98.0} & 96.4 & 99.7 & 98.4 & 93.8 & 100.0 & 99.0 & \textbf{99.4} & \textbf{99.5} & \textbf{99.5} & \textbf{97.9} \\
Multi-view PPT + 3DPE & 4 & 9.7G & \textbf{98.0} & \textbf{98.0} & 96.4 & 99.7 & \textbf{98.5} & \textbf{94.0} & \textbf{100.0} & \textbf{99.1} & 99.2 & 99.4 & 99.3 & \textbf{98.0} \\
\bottomrule
\end{tabular}
}
\caption{\small{2D pose estimation on Human3.6M. The metric is JDR on original image. All inputs are resized to $256\times256$. \#V means the number of views used in cross-view fusion step. The FLOPs is the total computation for each view and cros-view fusion. }}
\label{tab:h36M_2d}
\end{table}

\begin{table}[t]
% >>>>>>>>>>>>>>>>>>>>>>>>>>>>>>>>>>>>>>>>>>>> H36M MPJPE >>>>>>>>>>>>>>>>>>>>>>>>>>>>>>>>>>>>>>>>>>>>
\resizebox{1.0\textwidth}{!}{
\begin{tabular}{l|ccccccccccccccc|c}
\toprule
Method & Dir & Disc & Eat & Greet & Phone  & Pose & Purch  & Sit & SitD & Smoke & Photo & Wait & WalkD & Walk & WalkT & Avg \\
\midrule
% \midrule
Crossview Fusion\cite{qiu2019cross}  & 24.0 & 28.8 & 25.6 & 24.5 & 28.3 & 24.4  & 26.9 & 30.7 & 34.4  & 29.0   & 32.6 & 25.1 & 24.3 & 30.8 & 24.9 & 27.8 \\
Epipolar Trans. \cite{he2020epipolar} & 23.2 & 27.1  & 23.4  & 22.4  & 32.4  & \textbf{21.4} &  \textbf{22.6} & 37.3 & 35.4 & 29.0  & 27.7  & 24.2    & 21.2 & 26.6 & \textbf{22.3}  & 27.1   \\
TransFusion \cite{ma2021transfusion} & 24.4 & \textbf{26.4}  & 23.4 & \textbf{21.1} & 25.2 & 23.2 & 24.7 & 33.8  & \textbf{29.8} & 26.4 & 26.8 & 24.2 & 23.2 & 26.1 & 23.3 & 25.8 \\
\hline 
Multi-PPT+3DPE & \textbf{21.8} & 26.5 & \textbf{21.0} & 22.4 & \textbf{23.7} & 23.1 & 23.2 & \textbf{27.9}  & 30.7 & \textbf{24.6} & \textbf{26.7} & \textbf{23.3} & \textbf{21.2}  & \textbf{25.3} & 22.6 &  \textbf{24.4} \\
\bottomrule
\end{tabular}
}
\caption{\small{The MPJPE of each pose sequence on Human 3.6M.}}
\label{tab:h36M_3d}
\end{table}

\vspace{-0.5em}
\subsection{Results}
\vspace{-0.5em}
The 2D results on Human 3.6m is shown in Table \ref{tab:h36M_2d}. The MACs (multiply-add operations) consider both single-view forward MACs of all views and cross-view fusion MACs. 
Noticeably, our multi-view PPT outperforms all previous cross-view fusion methods on JDR.
% >> 3D PE
The JDR can be further improved with the 3D positional encodings (3DPE) \cite{ma2021transfusion} on visual tokens. 
Meanwhile, it can significantly reduce the computation of all 4 view fusion, \ie the MACs is reduced from $55.1$G to $9.7$G. 
% >>> vs other 2-view fusion methods
When only fusing $2$ views, multi-view PPT still achieves comparable accuracy with other two-view-fusion methods \cite{he2020epipolar,ma2021transfusion}, 
% >>> pruning vs no pruning 
Moreover, we add the baseline that adds transformers on top of TokenPose to perform cross-view fusion, which can be considered as multi-view PPT without token pruning.  
The JDR is $97.4\%$ (-$0.7\%$ with respect to our multi-view PPT), which supports that our human area fusion is better than global attention in both accuracy and efficiency. 
% 
% >>>>>>>>>>> 3D MPJPE >>>>>>>>>>>
The MPJPE of estimated 3D pose is reported in Table \ref{tab:h36M_3d}. We can observe that multi-view PPT also achieves the best MPJPE on 3D pose, especially on sophisticated action sequences such as ``Phone" and ``Smoke", as the result of 3D pose is determined by the accuracy of 2D pose. 
Therefore, our ``human area fusion" strategy is better than previous fusion strategies as it strikes a good balance between efficiency and accuracy. 
% 
% >>>>>>>>>>>> Ski Pose >>>>>>>>
We can also observe consistent conclusion on Ski-Pose from Table \ref{tab:skipose}. Nevertheless, it seems that the performance in this datatset tends to be saturated.  
The reason might be that there is limited number of training examples, thus the transformer is  easy to overfit.

\begin{table}[t]
% >>>>>>>>>>>>>>>>>>>>>>>>>>> SkiPose >>>>>>>>>>>>>>>>>>>>>>>>>>>
\centering
\resizebox{0.8\textwidth}{!}{
\begin{tabular}{l|c|c|c}
\toprule
Method  & MACs & 2D Pose / JDR (\%) $\uparrow$ & 3D Pose / MPJPE (mm) $\downarrow$ \\
\midrule
Simple Baseline-Res50 \cite{xiao2018simple}  & 77.6G  & 94.5  & 39.6  \\
TokenPose \cite{li2021tokenpose} & 16.8G & 95.0 & 35.6 \\
Epipolar Transformer \cite{he2020epipolar}   & 77.6G  & 94.9  & 34.2  \\
% TransFusion \cite{ma2021transfusion}         & 75.3G  & 96.0 & 31.6   \\
\textbf{Multi-view PPT} & \textbf{14.5G} &  \textbf{96.3}  &  \textbf{34.1} \\
\bottomrule
\end{tabular}
}
\caption{\small{2D and 3D pose estimation accuracy comparison on Ski-Pose.} }
\label{tab:skipose}
\end{table}

\vspace{-0.5em}
\subsection{Visualizations}
\vspace{-0.5em}

\subsubsection{Human Tokens. }
\begin{figure}[t]
    \centering
    \includegraphics[width=0.95 \linewidth]{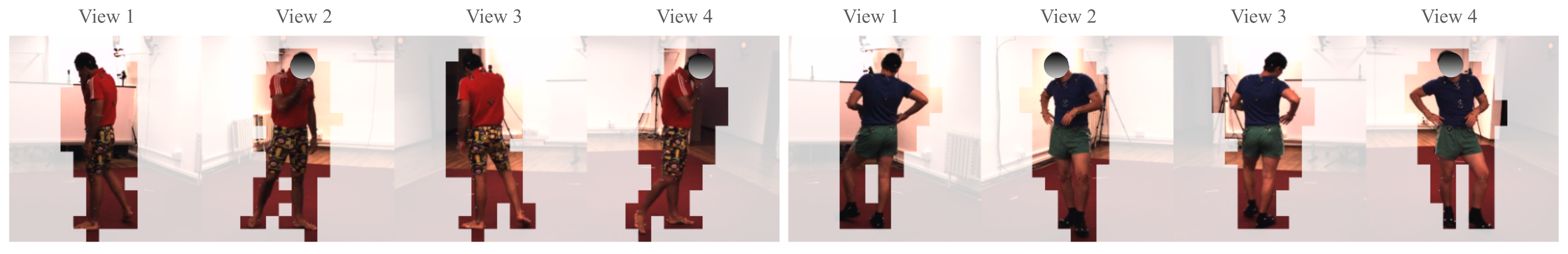}
    \caption{
    \small{Visualizations of the final located tokens on Human 3.6M validation set. For each group, each column is an image from one view. The masked regions represent the pruned tokens. We perform cross-view fusion among these selected tokens. }
    }
    \label{fig:human_mask_multi}
\end{figure}

Fig.\ref{fig:human_mask_multi} presents the selected human tokens in all views. 
%As consistent with
Similar to the conclusion on COCO, our PPT accurately locates all human areas and prunes background areas in all views. 
%To this end
Moreover, the tokens used in the cross-view fusion step can be significantly reduced.

\vspace{-0.5em}
\subsubsection{Qualitative results. }
\begin{figure}[t]
    \centering
    \includegraphics[width=0.88\linewidth]{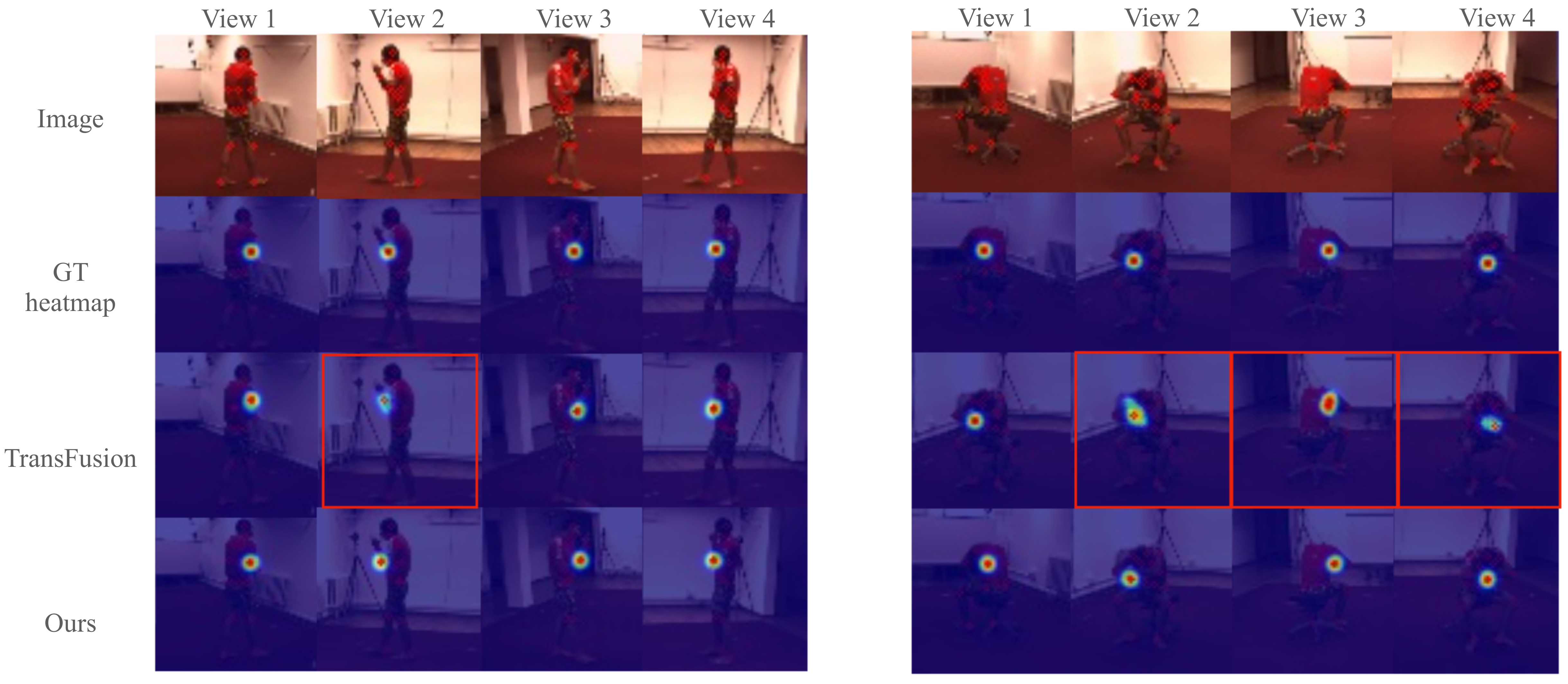}
    \caption{\small{Sample heatmaps of our approach.} }
    \label{fig:heatmap_sample}
\end{figure}

We present examples of predicted 2D heatmaps on the image in Fig.\ref{fig:heatmap_sample}, and compare our methods with TransFusion \cite{ma2021transfusion}. It is observed that our method can solve heavy occlusion cases very well, while TransFusion cannot.  
For two-view-fusion method, occlusion cases in current view may still be occluded in the neighbor view. For example, the heatmap marked with red box is inaccurate in both view 2 and view 4. Thus, fusing this bad quality heatmap cannot improve the final prediction. 
However, our method can avoid this problem by fusing clues from all views.

\begin{figure}[t]
    \centering
    \includegraphics[width=0.88\linewidth]{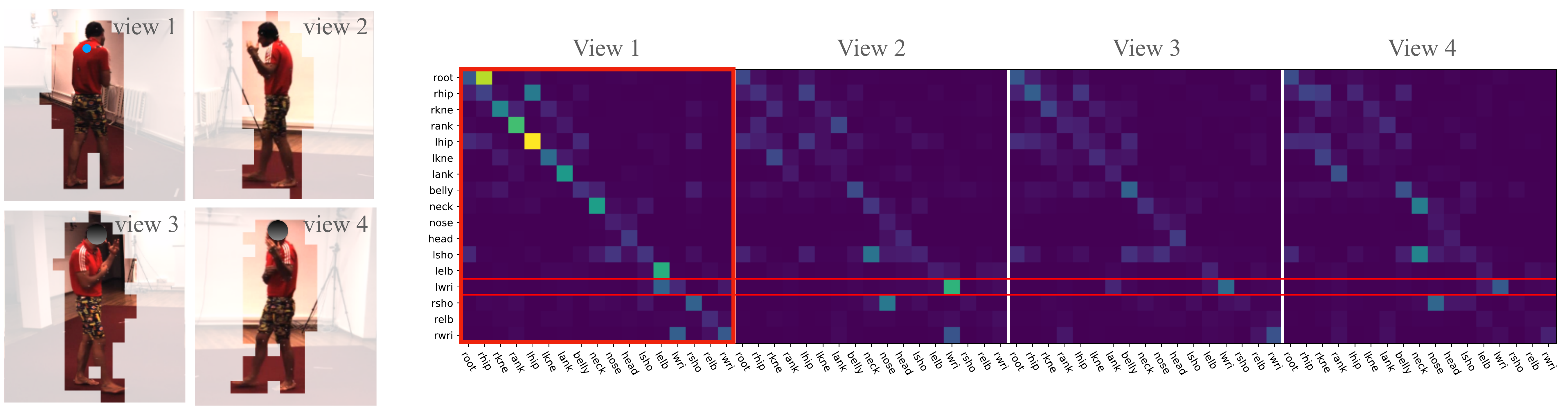}
    \caption{\small{Attention maps among keypoint tokens. } }
    \label{fig:attention_keypoint}
\end{figure}

\vspace{-0.5em}
\subsubsection{Attentions. }
We present an example of the attention map between keypoint tokens in Fig.\ref{fig:attention_keypoint}. 
Given keypoint tokens in one view, they pay attention to keypoints tokens in all views. For example, the left wrist in the first view (blue dot) is occluded, thus its corresponded keypoint token attends to the keypoint token in the second view, where the keypoint is visible.    
Therefore, the keypoint token in multi-view PPT can learn the dependencies among joints in different views.

\vspace{-0.5em}
\section{Conclusion}
\vspace{-0.5em}
In this paper, we propose the PPT for 2D human pose estimation. Experiments on COCO and MPII show that the PPT achieves similar accuracy compared with previous transformer-based networks but reduces the computation significantly. We also empirically show that PPT can locate a rough human mask as expected. 
Furthermore, we propose the multi-view PPT to perform the cross-view fusion among human areas. We demonstrate that multi-view PPT efficiently fuses cues from many views and outperforms previous cross-view fusion methods on Human 3.6M and Ski-Pose.

\clearpage
\bibliographystyle{splncs04}
\bibliography{egbib}

\newpage

\setcounter{section}{0}
\renewcommand\thesection{\Alph{section}}

\section{Appendix}
\subsection{Runtime evaluation for Monocular 2D pose estimation}
Although the GFLOPs reflects the efficiency of networks, it is not equivalent to the real runtime on hardware due to different implementation. 
We further report the throughput, which measures the maximal number of input instances the network can process in time a unit. 
Unlike FPS (frame per second), which involves the processing of a single instance, the throughput evaluates the processing of multiple instances in parallel. 
During the inference time of the top-down method, given one input image, multiple human instances located by an object detector are usually cropped, resized, and combined into a minibatch to accelerate the inference. Then the minibatch of multiple human instances is fed into the pose detector. 
Thus, we believe throughput is a more reasonable metric to evaluate top-down 2D human pose estimation networks.

\begin{figure}[t]
    \centering
    \includegraphics[width=0.99\linewidth]{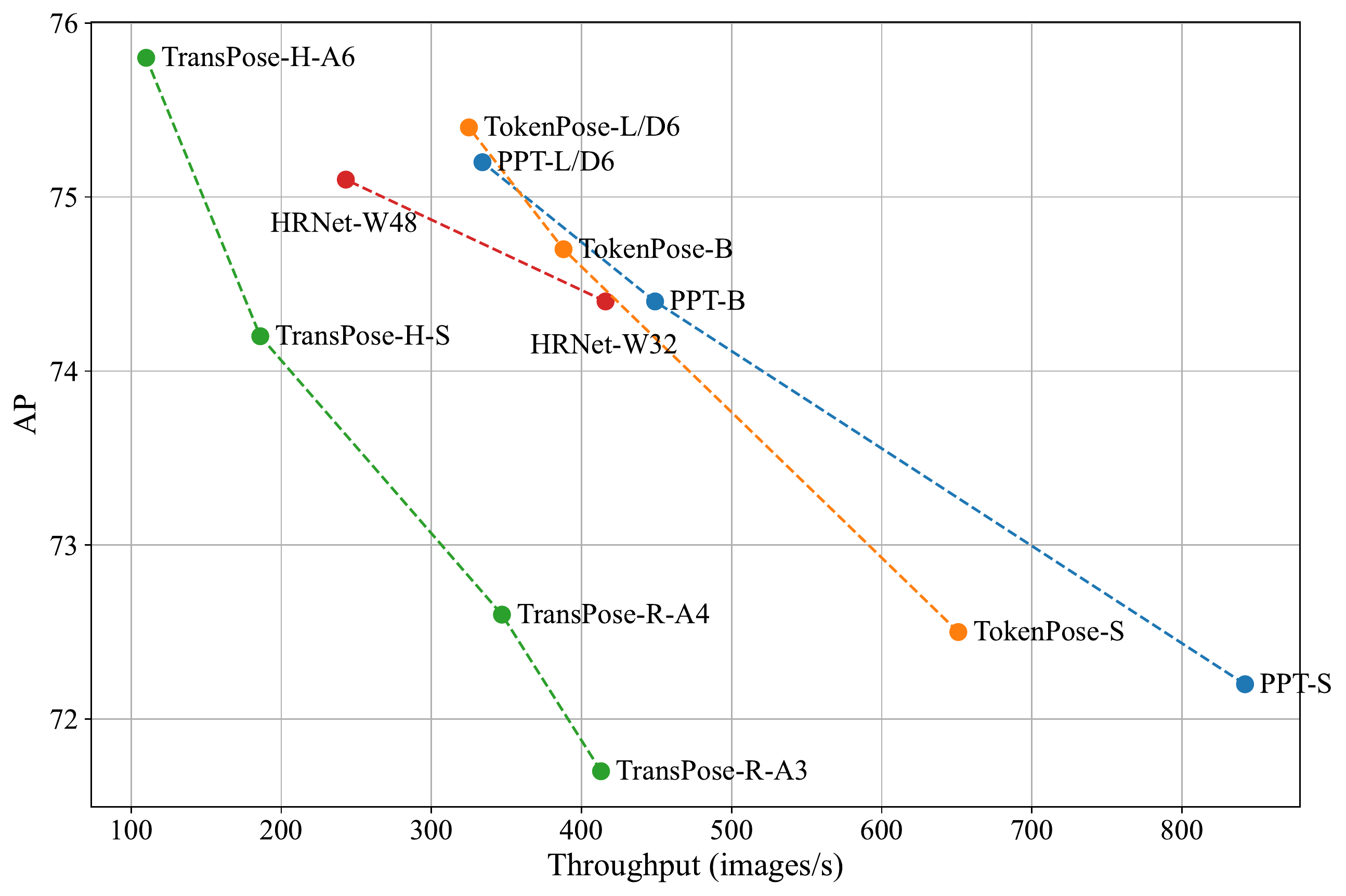}
    \caption{\small{Comparison of throughput on COCO validation dataset. }  }
    \label{fig:throughput}
\end{figure}

We set the batch size to $32$ for all networks, and compute the throughput on a single 2080 Ti GPU. Both FPS and throughput of PPT and TokenPose \cite{li2021tokenpose} are shown on Table \ref{tab:coco_fps}. 
Remarkably, pruning tokens cannot significantly improve the time of a single instance (\ie FPS). We believe the extra time introduced by the pruning operation is not negligible. 
Nevertheless, PPT significantly improves the throughput from TokenPose, which is consistent with the improvement of GFLOPs in Table \ref{tab:coco_val}. 
We further show the comparison of throughput with other methods in Figure \ref{fig:throughput}. Our PPT consistently improves the throughput at the same AP level. 
Thus, pruning token does improve the runtime on hardware in practice.

\begin{table}[h]
% >>>>>>>>>>>>>>>>>>>>>>>>>>> COCO >>>>>>>>>>>>>>>>>>>>>>>>>>>
\centering
\resizebox{0.6\textwidth}{!}{
\begin{tabular}{l|l|l|c|l}
\toprule
Method &  \#Params & AP  & FPS & Throughput  \\
\hline
TokenPose-S & 6.6M & 72.5 & 120 & 651 \\
PPT-S & 6.6M & 72.2 & 123 & 842 (+\textbf{30}\%) \\
\hline 
TokenPose-B & 13.5M & 74.7 & 50 & 388\\
PPT-B & 13.5M & 74.3 & 51 & 451 (+\textbf{16}\%) \\
\hline 
TokenPose-L/D6 & 20.8M & 75.4 & 60  & 325 \\
PPT-L/D6 & 20.8M & 75.2  & 61 & 334 (+\textbf{3}\%) \\

\bottomrule
\end{tabular}
}
\caption{ \small{FPS and Throughput on COCO validation dataset. } }
\label{tab:coco_fps}
\end{table}

% The TransPose \cite{yang2020transpose} maintains a high-resolution feature map with $1024$ tokens, making it difficult  

\end{document}